# AFS: An Attention-based mechanism for Supervised Feature Selection

Ning Gui[1], Danni Ge[2], Ziyin Hu[2]

[1]School of Software, Central South University, China, ninggui@gmail.com

[2] School of Informatics, Zhejiang Sci-Tech University, China, dongfangdecheng@gmail.com, fayssica@gmail.com

**Abstract**

As an effective data preprocessing step, feature selection has shown its effectiveness to prepare high-dimensional data for many machine learning tasks. The proliferation of high dimension and huge volume big data, however, has brought major challenges, e.g. computation complexity and stability on noisy data, upon existing feature-selection techniques. This paper introduces a novel neural network-based feature selection architecture, dubbed Attention-based Feature Selection (AFS). AFS consists of two detachable modules: an attention module for feature weight generation and a learning module for the problem modeling. The attention module formulates correlation problem among features and supervision target into a binary classification problem, supported by a shallow attention net for each feature. Feature weights are generated based on the distribution of respective feature selection patterns adjusted by backpropagation during the training process. The detachable structure allows existing off-the-shelf models to be directly reused, which allows for much less training time, demands for the training data and requirements for expertise. A hybrid initialization method is also introduced to boost the selection accuracy for datasets without enough samples for feature weight generation. Experimental results show that AFS achieves the best accuracy and stability in comparison to several state-of-art feature selection algorithms upon both MNIST, noisy MNIST and several datasets with small samples.

## Introduction

With the rapid advancement of Internet of Things and industrial automation systems, enterprises and industries are collecting and accumulating data with unparalleled speed and volume(Yin *et al.*, 2014). The large amount of data makes the data-driven modeling approach in many domains desirable with the automatic knowledge discovery. In order to extract useful information from huge amounts of otherwise meaningless data, one important machine-learning technique is feature selection (FS), which directly applies a subset of relevant features for the learning tasks. Those irrelevant, redundant and noisy features in respect to the supervision target are ignored. The simplified feature set often results in a more accurate model which is much easier to understand. Many different methods have been proposed and effectively used for various tasks.

In the era of big data, however, most off-the-shelf feature selection methods suffer major problems: varying from the computation scalability to the stability. For instance, many existing algorithms demand the whole dataset loaded into the memory before calculation, which becomes infeasible when data scales to terabytes. Furthermore, those datasets normally contain a lot of noisy/outlier samples. It is observed that many well-known feature selection algorithms suffer from the low stability problem after small data perturbation is introduced in the training set(Alelyani *et al.*, 2011). Our experience on one industrial dataset with 16K features and 16M records validates this conclusion as many existing solutions are incapable, slow or unstable upon this set. As pointed out in their review paper(Bolón-Canedo *et al.*, 2015): "it is evident that feature selection researchers need to adapt challenges posed by the explosion of big data."

Deep-learning-based feature selection methods (Wang *et al.*, 2014; Li *et al.*, 2015; Roy *et al.*, 2015; Zhao *et al.*, 2015) are considered to have the potential to cope with the "curse of dimensionality and volume" of big data because deep neural networks have been proved effective for processing massive data. Among many techniques proposed in deep learning, the attention mechanism, a recent proposed technique to focus on the most pertinent piece of information, rather than using all available information, has already gained much success in various machine learning tasks, e.g. natural language processing(Yin *et al.*, 2015) and image recognition(Xu *et al.*, 2015). Interestingly, the attention generation process is quite similar to the feature selection process as they both focus on selecting partial data from the high dimensional dataset. It becomes the initial inspiration of our work.

In this paper, a novel attention-based supervised feature selection architecture, called AFS, is proposed to evaluate feature attention weight (short as *feature weight*, interchangeable with the term *feature score* typically used in feature selection) by formulating the correlation problem among features and supervision target into a binary classification problem, supported by a shallow attention net for each feature. This architecture is able to generate attention weights for both classification and regression feature selection problems. The main contributions of our work are as follows.

- **A novel attention-based supervised feature selection architecture:** the architecture consists of an attention-based feature weight generation module and a learning module. The detachable design allows different modules to be individually trained or initialized.
- **An attention-based feature weight generation mechanism:** this mechanism innovatively formulates the feature weight generation problem into a feature selection pattern problem solvable with attention mechanism.
- **A model reuse mechanism for computation optimization:** is proposed that can directly reuse existing models to effectively reduce the computation complexity in generating feature weights.
- **A hybrid initialization method for small datasets:** is proposed to integrate existing feature selection methods for weight initialization. This design extends AFS's usage to small datasets in which AFS might not have enough data for feature weight generation.

A set of experiments are designed on both Large-dimensionality Small-instance dataset (denoted as L/S *dataset*) and Medium/large-dimensionality Large-instance dataset (short for M/L dataset). The highest feature selection accuracy and moderate computation overhead, compared with existing baseline algorithms, have been observed on both the MNIST dataset and the challenging noisy MNIST (n-MNIST). The proposed model reuse mechanism can compute the attention weights about 10 times faster with similar accuracy. The hybrid initiation method can also boost the classification accuracy from 1.09% to 6.61% upon the Relief and Fisher Score methods on two tested L/S datasets. To the best of our knowledge, AFS is the first attention-based neural network solution for general supervised feature selection tasks.

# Related Work

This section first reviews the state-of-the-art supervised feature selection works. Then researches in the attention mechanism domain are illustrated.

## Feature Selection methods

The supervised feature selection methods are normally categorized as wrapper, filter, and embedded methods(Bengio *et al.*, 2003; Gui *et al.*, 2017).

The wrapper methods rely on the predictive accuracy of a predefined learning algorithm to evaluate the quality of selected features. They generally suffer the problem of high computation complexity(Tang *et al.*, 2014). The filter methods separate feature selection from learning algorithms and only rely on the measures of the general characteristics of the training data to evaluate the feature weights. Different feature selection algorithms exploit various types of criteria to define the relevance of features: e.g. similarity-based methods, e.g. SPEC(Zhao and Liu, 2007) and Fisher score(Duda *et al.*, 2012), feature discriminative capability, e.g. ReliefF(Robnik-Šikonja and Kononenko, 2003), information-theory based methods, e.g. mRmR(Peng *et al.*, 2005) and statistics-based methods ,e.g. T-Score(Shumway, 1987). The embedded methods depend on the interactions with the learning algorithm and evaluate feature sets according to the interactions. Normally, appropriate regularizations are added to make the certain feature weights as small as possible to facilitate convergence, e.g. FS with l2,1-Norm(Liu *et al.*, 2009).

In order to handle the computation complexity of big data, limited deep-neural-network based methods have been proposed. Li et al.(2015) proposed a deep feature selection (DFS) by adding a sparse one-to-one linear layer. As the network weights are directly used as the feature weights, it cannot handle situations where inputs have outliers or noise. Towards this end, Roy et al.(2015) use the activation potentials contributed by each of the individual input dimensions, as the metric for feature selection. However, this work relies on the specific DNN structure and the ReLU activation function which might not be so suitable in many learning tasks.

Recent trends of feature selection methods are more focused on data with specific structures, e.g. distributive fairness(Grgic-Hlaca *et al.*, 2018), multi-source data(Liu *et al.*, 2016) or streaming data(Zhang *et al.*, 2015). However, we argued, their work still largely relies on the advances of feature selection methods on conventional data.

## Attention in Neural Network

The attention mechanism is a method that takes arguments, and a context and returns a vector supposed to be the summary of the arguments, focusing on information linked to the context. It has been successfully used first in visual image domain and then extended to various fields, e.g. language translation and audio processing tasks. Normally, the attention-based methods are applied to data with specific structures, e.g. spatial, temporal or mixture of spatial and temporal structures.

In respect to the inputs with a spatial structure (such as a picture), the construction of attention focuses on the salient part of image. The work by Girshick et al.(2014) uses a region proposal algorithm, and Erthan et al.(2014) show that it is possible to regress salient regions with a CNN. Jaderberg et al.(2015) proposed the spatial transformer networks by using bilinear interpolation to smoothly crop the activations in each region. In order to acquire POI in the image, Laskar et al.(2017) maintain all features within the salient region but weakens the values of the background region.

For inputs with a temporal structure (i.e. language, video), the attention mechanism is used to obtain the relationship between current inputs and previous inputs, by Recurrent Neural Networks such as RNN or LSTM. For example, Yao et al.(2015) use the LSTM to extract latent representation of video. Ma et al.(2017) use the RNN to encode the Patient HER data which consist of sequences of visits over time. One peculiar trait of temporal data is the correlations can be furthered divided into two levels: local attention for localized correlation and global attention for more remote correlation. In order to integrate different correlation, the hierarchical structure is adopted (Tong *et al.*, 2017).

Those above discussed researches normally provide domain-specific attention-based solutions for data with a certain structure. However, in many feature selection tasks, the data structure is not so obvious or is hard to obtain. In this paper, we focus on the feature selection for conventional data without any pre-knowledge.

## AFS Architecture

In this section, the overall architecture of AFS is illustrated and analyzed. Two extensions, namely the model reuse mechanism and the hybrid initialization are introduced.

### Notation

This paper presents matrix as uppercase character (e.g. A), and vector as lowercase (e.g. a). For example, a dataset is presented by a matrix $X = \{X_i^k | i = 1,2,\ldots,m; k = 1,2,\cdots,d\} \in R^{d \times m}$, where m is the number of samples, and d is the number of features. $X^T$ is used to denote the transpose of matrix X. Each sample is denoted by a column vector $x_i$, $i = 1,2,\cdots,m$, each feature is denoted by row vector $x^k, i = 1,2,\cdots,d$, and the k-th feature of $x_i$ is denoted by $x_i^k$. $x_i$ is associated with the label $y_i$. For multi-class task, $y_i^j$ presents the label belongs to j-th class.

### Architectural design

Similar to the embedded methods, our proposed AFS architecture embeds feature selection with learner construction process. As shown in the Figure1, AFS consists of two major modules, namely, the **attention module** and the **learning module**. The attention module is on the upper part of AFS and is responsible for computing the weights for all features. As shown in this figure, the attention module is the core of the whole framework. The learning module aims to find the optimal correlation between the weighted features and the supervision target by solving the optimization problem. It connects the supervision target and features by the back propagation mechanism, and continuously corrects the feature weights during the training process. The attention module and the learning module, together build the correlation that best describes the degree of relevance of the target and features.

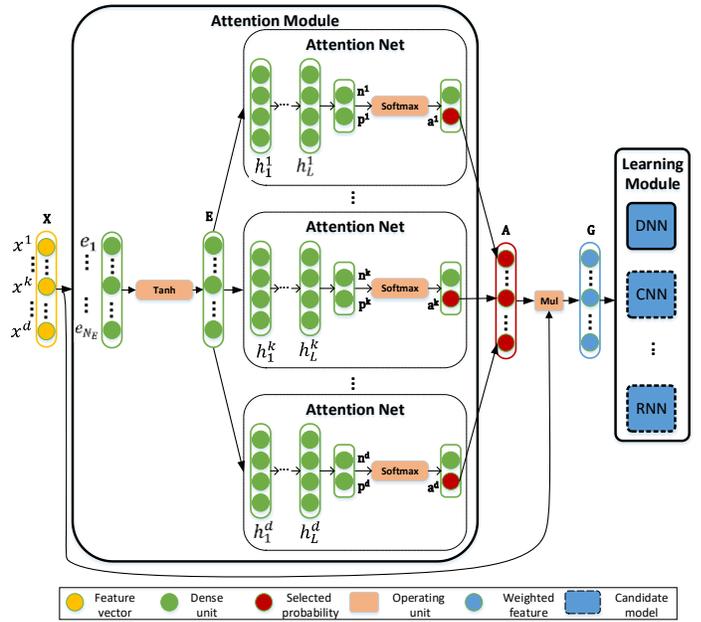

*Figure 1. The detachable architecture of AFS, with an attention module and a replaceable learning module*

AFS is designed with a loosely coupled structure. Both the attention module and the learning module can be individually customized to match a specific task, especially for the learning module. Currently, deep learning communities have generated thousands of off-the-shelf models which can be directly reused by AFS as the learning module. The parameters of the attention module can then be generated with much lower computation overhead. In addition, for L/S datasets, AFS might not have enough samples to train the network. In order to solve this problem, a hybrid initialization method is proposed which uses existing feature selection algorithms to initialize the weights of the attention module.

### Attention Module

In order to represent the correlation between the features and the supervision target, we convert the correlation problem into a binary classification problem: for a specific supervision label, whether a feature should be selected. Then, the feature weights are generated according to the distribution of the feature selection pattern.

Firstly, a dense network is used to extract the intrinsic relationship (denoted as E) among the raw input X and eliminate certain noise or outliers.

$$E = \text{Tanh}(X^T W_1 + b_1) \quad (1)$$

where $W_1 \in R^{d \times N_E}$ and $b_1 \in R^{N_E}$ are the parameters to be learned. The introduced dense network E compresses the original feature domain into a vector with a smaller size (adjustable according to specific problems), while keeps the major part of the information. As the size of E is normally much smaller than the size of features, certain redundant features will be discarded during this process. The design also reduces impacts from individual noises. The units of the dense neural network $N_E$ could be adjusted according to respective tasks. Here, the dense network is chosen as no structural assumption of the inputs are made. For structural data, it can be replaced with other types of neural networks. The nonlinear function $\text{Tanh}(\cdot)$ is adopted as it retains both positive and negative values. Therefore, it can preserves important information during the extraction of E.

Secondly, by using the extracted E as input, each feature $x^k$ is assigned with a shallow neural network to determine its probability of selection, namely *attention net*. In this net, we do not directly use the typical soft attention mechanism in which softmax action is used to generate weighted arithmetic mean of all the feature values. This action will result in relatively small weights for most features, and only a very small number of features with relatively large weights. It is good for feature division while suffers the loss of details on the whole feature sets. Furthermore, when there are outliers or noise in the data, some extraneous features might also erroneously be given a certain big weight.

In this paper, a different soft-attention mechanism is proposed. Due to the binary classification (select/unselected), each attention unit in the attention layer generate two values: for the k-th feature, $p^k, n^k$ represent selected/unselected values respectively, calculated with Eq. (2) and (3):

$$p^k = w_p^k h_L^k + b_p^k \qquad (2)$$
$$n^k = w_n^k h_L^k + b_n^k \qquad (3)$$

where $h_L^k$ is the output of L-th hidden layer in the k-th attention net. $w_p^k, b_p^k$ are the parameters of $p^k$ and $w_n^k, b_n^k$ are the parameters of $n^k$ to be learned. Due to the fact that $p^k$ and $n^k$ might be quite close. The softmax is then used to generate differentiable results to statistically boost the difference between selection and un-selection, with each possibility is in the range (0, 1). Here, we only focus on $p^k$ as it generates the probability of being selected as attention feature $a^k$ as follows:

$$a^k = \frac{\exp(p^k)}{\exp(p^k) + \exp(n^k)} \qquad (4)$$

Those shallow attention nets generate attention matrix $A = \{a_i^k | i = 1,2,\cdots,m; k = 1,2,\cdots,d\} \in R^{d \times m}$. According to the attention matrix, the weight of k-th feature is calculated with $s^k = \frac{1}{m}\sum_{i=1}^{m} a_i^k$. Note that the parameters of attention module are summarized as $\theta_a$.

**Discussion:** Compared with the embedded feature selection method, the attention module has more obvious advantages: 1) the feature weights are generated by the feature selection pattern, produced via separated attention networks, rather than coefficient values adjusted only by backpropagation. The intrinsic relationship between the features can be more comprehensively considered by the neural network E; 2) the feature weights are always limited to a value between 0 and 1, which can accelerate training convergence. Furthermore, the softmax design is a fully differentiable deterministic mechanism that is easy to train with backpropagation. 3) redundant features are removed with the joined work from both E and the learning module. Due to the smaller size of E certain information of redundant features will be discarded. Then, the attention net corresponding those discarded features might not have enough information and generates a low feature weight for these features. Thereby, the output of the redundant features can be further suppressed. Of course, which redundant features are to discard is highly random.

### Learning module

By using the pair-wised multiplication $\odot$ to contact the feature vectors X and A, we obtain the weighted features G as follows:

$$G = X \odot A \qquad (5)$$

The process of constantly adjusting the A is equivalent to making a trade-off between selecting and un-selecting. In order to generate an attention matrix A, the learning module runs with backpropagation by solving the objection function as follows:

$$\arg\min_A loss \left[f_{\theta_l}\left(G_{\theta_a}(X)\right) - Y\right] + \lambda R(\theta) \qquad (6)$$

where $\theta = \langle \theta_a, \theta_l \rangle$ and $R(\cdot)$ is often an L2-norm that helps to speed up the optimization process and prevent overfitting. Here, $\lambda$ controls the strength of regularization. The loss function depends on the type of prediction tasks. For classification tasks, the cross entropy loss functions are usually used. For regression tasks, the mean-square error (MSE) is normally used. Note that $f_{\theta_l}(\cdot)$ is a neural network with parameters $\theta_l$.

For a specific learning problem, AFS can use a network structure that best fits for the particular task. Currently supported network structures include: e.g. deep neural networks (DNN), convolutional neural networks (CNN), and Recurrent Neural Network (RNN).

### Learning model reuse mechanism

As shown in Figure 1, the computational complexity of AFS comes from both the attention module and the learning module. For many learning tasks, there already exists a large number of dedicated trained models, e.g. ResNet and VGG. Since the AFS structure is detachable, the training of the learning module part and the attention module part can be learned separately. This design allows existing models directly to be reused in AFS.

Using the trained model, the saved model weights are initialized to the model parameters portion of AFS, called

**AFS-R**. Since the parameters in the learning module are already converged, only a few tunes are needed. There are 2 ways to train the AFS-R**:** fine-tuning the both attention module and learning module (denoted as AFS-R-GlobalTune**)** or fixing the learning module and only train the attention module **(**denoted as AFS-R-LocalTune**)**. The first way is to train the attention module and the learning module at the same time. The second way is to fix the learning module and only train the attention module.

### A hybrid initialization method

Since the AFS benchmark trainer is a neural network, its performance highly depends on the number of samples in the dataset. Thus, a small number of samples might not be able to generate enough propagation to tune the whole neural network. In order to extend the usability of AFS on small datasets, a hybrid initialization method is proposed by reusing certain feature selection method's results as the initial feature weights.

This hybrid initialization method can be divided into three major steps: 1) Generating the feature weights $W_{fs}$ with a certain feature method. Since it might have value ranges other than the possibility range [0,1], Min-Max normalization is used to normalize $W_{fs}$; 2) Pre-training *the attention module*. In this step, each sample is tagged with $W_{fs}$ instead of the original label. We only train the attention module using the constructed dataset. Note that the objective function of this part is Eq. (7). To optimize the objective function, we employ adaptive moment estimation (Adam) for optimizing the attention module; 3) Training the AFS with the normal training process. Note that in this process, the objective function of this part is Eq. (6) instead of Eq. (7).

The second step in this algorithm is a regression task which aims to initialize the attention matrix A to match $W_{fs}$. Therefore, the objective function adopts the MSE as loss function as follows:

$$arg \min_{\theta_a} \frac{1}{m} \sum_{i=1}^{m}(a_i - W_{fs})^2 + \lambda R(\theta_a) \qquad (7)$$

## Experiments

In this section, we will conduct experiments to answer the following research questions:
- **Q1** Does AFS outperform state-of-the-art feature selection methods?
- **Q2** How much computational complexity can be reduced by reusing existing models?
- **Q3** How to use the existing feature selection method combined with AFS to improve the feature selection performance upon L/S datasets?

We has open-sourced AFS and the source code can be found at https://github.com/upup123/AAAI-2019-AFS.

---

[1] http://yann.lecun.com/exdb/mnist/
[2] http://csc.lsu.edu/~saikat/n-mnist/

### Experiment Settings

**Datasets**. The datasets used for experiments are summarized in Table 1. The MNIST dataset[1] consists of grey-scale thumbnails, 28 x 28 pixels, of handwritten digits 0 to 9. The n-MNIST dataset[2] consists of three MNIST variants with: 1) white Gaussian noise (denoted as n-MNIST-AWGN); 2) motion blur (denoted as n-MNIST-MB); 3) a combination of additive reduced contrast and white Gaussian noise (denoted as n-MNIST-RCAWGN). These datasets are selected due to the facts that MNIST has been intensively investigated and the n-MNIST dataset provides a good foundation for feature selection stability evaluation. The Lung_discrete and Isolet[3] are L/S datasets.

*Table 1 Datasets information*

| Datasets | Type | Classes | Samples | Features |
|---|---|---|---|---|
| MNIST | M/L | 10 | 70000 | 784 |
| n-MNIST-AWGN | M/L | 10 | 70000 | 784 |
| n-MNIST-MB | M/L | 10 | 70000 | 784 |
| n-MNIST-RCAWGN | M/L | 10 | 70000 | 784 |
| Lung_discrete | L/S | 7 | 73 | 325 |
| Isolet | L/S | 26 | 1560 | 617 |

**Evaluation Protocols:** The feature weights are obtained through the training data, and then they are sorted, and a certain number of features are selected as a feature subset in descending order. The accuracy of the feature subset on the test set is used as the performance metric. For L/S datasets, 3 times 3 fold cross-validation is adopted to provide a fair comparison.

**Baselines.** After the evaluation of the computational cost of existing feature selection methods on M/L datasets. RFS (Robust Feature Selection)(Nie *et al.*, 2010) and Trace ratio criterion (Nie *et al.*, 2008) are discarded as their execution time on MNIST exceeds15 hours**.** The performance of AFS is compared with the following feature selection methods. Unless explicitly stated, implementations of those algorithms are from the scikit-feature selection repository[3].

**Filter-Based Methods:**
- Fisher Score(He *et al.*, 2006) selects features according to their similarities.
- ReliefF (Kononenko, 1994) selects features by finding the near-hit and near-miss instances using the l1-norm.

**Embedded methods:**
- FS_l21(Feature selection with l2,1-norm)(Liu *et al.*, 2009) uses l2,1-norm regularization which is convex similarly to l1-norm regularization.
- RF(Random Forest)**,** a tree-based feature selection method provided by scikit-learn package.
- Roy *et al.* (2015): a DNN-based feature selection method, reproduced according to the paper

---

[3] http://featureselection.asu.edu

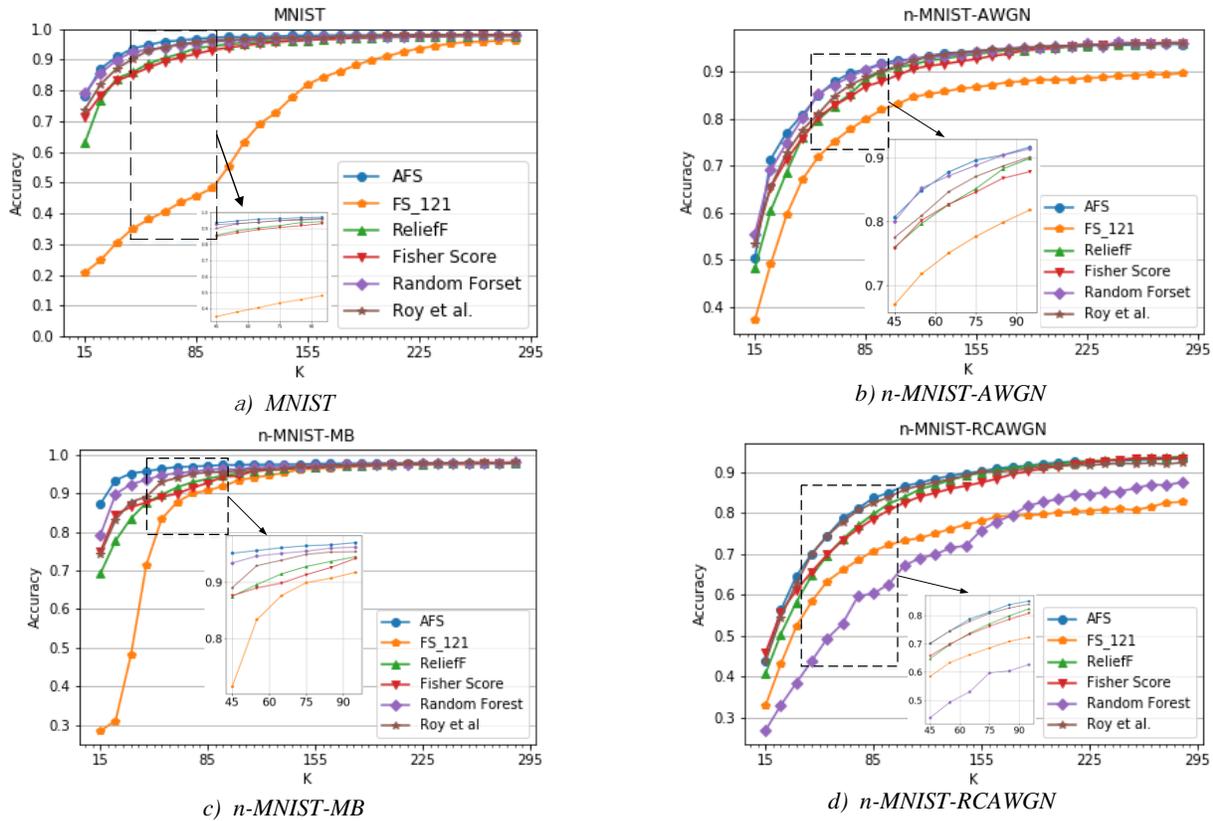

a) MNIST

b) n-MNIST-AWGN

c) n-MNIST-MB

d) n-MNIST-RCAWGN

Figure 2: Comparisons among different feature selection methods for the MNIST and n-MNIST datasets. The numbers of selected features are set from 15 to 295 with an interval of 10.w.r.t K denotes the number of features selected, results for AFS is after 3000 training steps.

**Parameter Settings.** Model parameters are initialized with truncated normal distribution with a mean of 0 and standard deviation of 0.1. The model is optimized by Adam, with batch size 100. The weight of regularizer is 0.0001. For the solution of AFS and Roy, the training step is both set to 3000 as they both begin to converge.

### Experiments on MNIST variants (Q1)

In order to have a fair comparison towards MNIST and its variants, the performance of major stream of classifiers are tested, with respect to the modeling accuracy.

**Classifier Selection**

Five different classifiers are tested: Decision Tree (DT), Gaussian Naïve Bayes(GNB), Random forest(RF), linear Support Vector Machine (SVM) and Neural Network(NN, here one hidden layer with 500 neurons is used). The results are shown in Figure 3. As SVM needs more than 2 hours for one round of experiment, its results are discarded. As shown in this figure, NN achieve the best modeling accuracy in all four datasets. NN achieves about 98% maximum modeling accuracy in the MNIST, much better than others. It also has the most stable performance. For all four datasets, NN achieves more than 94% maximum accuracy. In contrast, RF achieves goods results in the MNIST and n-MNIST-MB datasets while comparably poor results in the other two. Furthermore, GNB reaches max accuracy when K is 155 rather than around 295 in MNIST. Those classifiers are incapable for fair comparisons. Thus, NN is chosen as the default classifier and the comparisons of accuracy are based on the modeling results of the NN classifier.

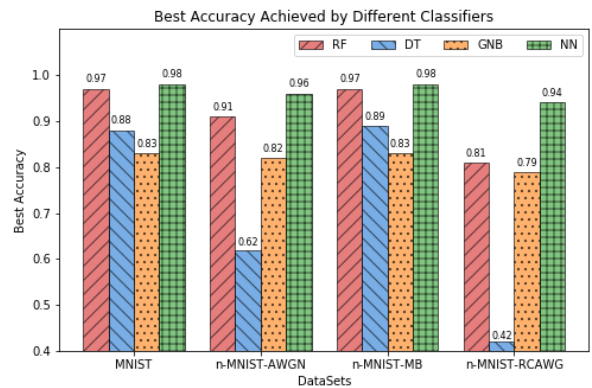

Figure 3: Comparisons among different classifiers w.r.t modeling accuracy

**Experiment Results**

The feature weights from different methods are sorted according to the numerical value (including positive and negative values, wherein negative values do not indicate negative correlation), respectively select the first K features after sorting, put them into the benchmark classifier, and fit with train data. For all these methods, we report the average results in terms of classification accuracy.

Figure 2 shows the modeling accuracy from different feature selection methods, with respect to MNIST and its variants. From both Figure 2 and Table 2, we can observe that:

1) AFS achieves the best accuracy on all four datasets and on almost all feature selection ranges. It significantly outperforms the compared methods. As shown in Table 2, AFS archives about to 3%~9% absolute accuracy improvements towards Fisher score and ReliefF methods. For the RF and Roy et al. solutions, AFS still maintains clear advantages over MNIST, n-MNIST-AWGN and n-MNIST-MB. For the n-MNIST-RCAWGN dataset, Roy et al. and AFS achieve comparative similar performance, with AFS leading Roy method for about 0.8%.

2) AFS achieves the best feature selection stability with respect to different types of noise. As shown in both Figure 2b, 2c and 2d, no matter what kind of noise is introduced, AFS exhibits almost consistent good performance. The solution of Roy achieves good average accuracy on the n-MNIST-RCAWGN dataset, while not as good performance on the other three datasets. The same happens to the RF method, which is quite sensitive to the reduced contrast noise and suffers almost the lowest accuracy on the n-MNIST-RCAWGN dataset. The two well-accepted ReliefF and Fisher Score have comparably stable performance. However, their accuracy is not so well.

3) As shown in Table 2, AFS outperforms other five methods significantly when the number of selected features is within the range of 15 to 85. This result shows that AFS has the most accurate feature weight ordering which is an important advantage for many modeling processes.

*Table 2. Average Accuracy (top15~top85 features), methods have the most significant accuracy improvement within this range*

|  | MNIST | n-MNIST-AWGN | n-MNIST-MB | n-MNIST-RCAWGN |
|---|---|---|---|---|
| AFS | **91.65** | **79.00** | **94.78** | **69.08** |
| FS_l21 | 34.81 | 64.64 | 66.34 | 56.90 |
| ReliefF | 84.28 | 73.60 | 85.67 | 64.24 |
| Fisher Score | 84.65 | 75.19 | 87.02 | 65.76 |
| RF | 90.54 | 78.86 | 91.99 | 45.53 |
| Roy et al. | 88.83 | 76.29 | 88.91 | 68.28 |

**Feature weight Visualization**

In order to further explore the reason why different methods display significant performance variations, the distribution map of the top selected features towards MNIST and its variants are shown in Figure 4.

Figure 4 shows that many top65 features selected by FS-121 are on the edge of the image, while the handwritten digits are almost located in the center. It explains the poor performance of FS-121. For the RF and Roy, they display totally different feature distribution patterns towards different datasets. It shows that they are to some extent impacted by the introduced noise. RF responses poorly to the reduced contrast noise. Roy fails to capture the features in the lower part of images in the MNIST dataset which is important for many handwritten digits, e.g. 5, 6, 8. It explains Roy's poor performance on MNIST dataset. The relatively stable Fisher score and ReliefF, like the AFS method, focus on the central area of the image. As the feature selection scale increases, from top65 to top350, the selected area of the image gradually expands to the periphery. The TOP65 feature selection area of AFS is divided into two areas, which accurately represents the structure of handwritten digits. Meanwhile, the distribution of TOP65 of ReliefF and Fisher are too concentrated to identify most important features.

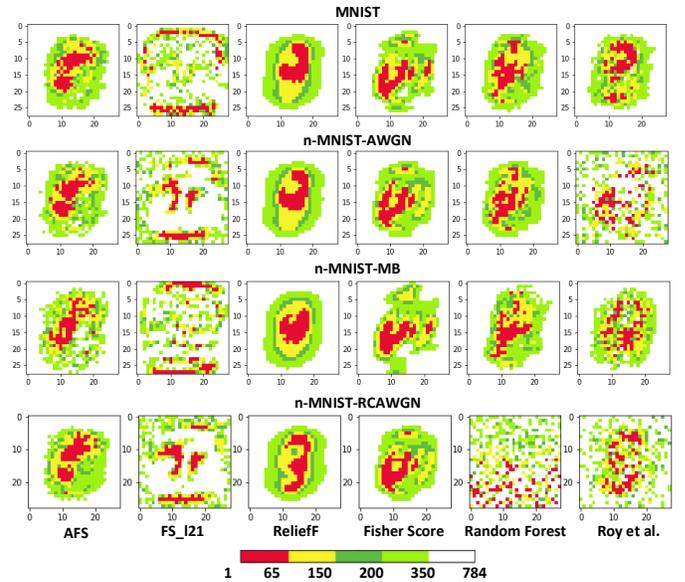

*Figure 4. Feature distribution map of five feature selection methods: dots with different color represent different sets of selected features.*

**Computational complexity**

In Table 3, the computation overheads of different feature selection methods are illustrated.

*Table 3. Comparisons of the computation overhead (in seconds)*

|  | MNIST | n-MNIST-AWGN | n-MNIST-MB | n-MNIST-RCAWGN |
|---|---|---|---|---|
| AFS(3000) | 218 | 201 | 190 | 199 |
| FS_l21 | 491 | 25 | 68 | 27 |
| ReliefF | 8557 | 7867 | 8159 | 7927 |
| Fisher Score | 586 | 509 | 517 | 527 |
| RF | 8 | 23 | 14 | 23 |
| Roy et al. | 10 | 11 | 11 | 12 |

The overhead is measured with the execution time for the feature weight generation process. Results show that AFS has moderate computation complexity. For the training with 3000 steps, it takes about 190 to 218s for the feature weight generation. The statistics-based methods, Fisher Score and ReliefF suffer the high computation cost. However, they are by no means the worst algorithms in terms of computation overhead as several algorithms, e.g. RFS and trace-ratio fail to calculate results on this dataset within 15 hours. RF and

Roy et al methods have very low computation overheads as they are embedded solutions and have fewer parameters to be tuned than AFS.

## Model reuse (Q2)

To evaluate the contribution of reusing existing models to reduce computation complexity, we directly use a DNN model that archives 98.4% classification accuracy on the MNIST test set(with 30000 training steps and 84s training time) as the learning model. Both AFS-R-GlobalTune and AFS-R-LocalTune strategies are tested.

*Table 4 Accuracy of reuse strategies, K is feature selected.*

|  | steps | Times(s) | Top K features selected | | | |
|---|---|---|---|---|---|---|
|  |  |  | K=25 | K=35 | K=65 | K=95 |
| AFS | 3000 | 218 | 87.05 | 90.91 | 95.84 | 97.06 |
| AFS-R-GlobalTune | 10 | 21 | 63.68 | 73.03 | 89.01 | 92.94 |
|  | 40 | 25 | 82.09 | 88.02 | 94.50 | 95.76 |
| AFS-R-LocalTune | 10 | 19 | 72.33 | 81.52 | 93.37 | 94.89 |
|  | 40 | 23 | 76.59 | 85.21 | 91.17 | 95.40 |

As seen in Table 4, when the tuning step is small (e.g. 10), the AFS-R-LocalTune solution normally achieves better accuracy than the AFS-R-GlobalTune as it has much fewer parameters to be tuned. When the tuning step increases, AFS-R-GlobalTune gradually achieves better accuracy. After 40 steps, it outpaces the AFS-R-LocalTune solution with much higher accuracy in all compared features sets, as the reused model has been better trained. Its results are quite close to those of AFS(3000). However, the computation overhead of AFS-R-GlobalTune-40 is about 25s, about 11.5% used by AFS (3000). For the same step, the global tune takes a little more time than the local tune solution as it has more parameters to be tuned.

## Performance on L/S datasets (Q3)

According to the hybrid initialization method, Fisher score and ReliefF are used as the base for initialization. These extended AFS are denoted as AFS-Fisher and AFS-ReliefF. Figure 5a and 5b show the experimental results on two L/S datasets. The SVM-linear classifier, with normally good performance on L/S datasets, is adopted.

As shown in Figure 5a, in Lung_discrete, AFS-Fisher achieves 81.81% average accuracy, 1.09% above its peer. AFS-ReliefF achieves 83.17% average accuracy, 1.73% above ReliefF. In Isolet, as shown in Figure 5b, AFS-ReliefF achieves 75.51% average accuracy, about 1.77% accuracy improvement than ReliefF. AFS-Fisher achieves the highest average accuracy 79.38%, well above other solutions. It shows that the hybrid initialization method can significantly improve existing feature selection algorithms' accuracy. Furthermore, one observation is that a higher sample/feature ratio might help the hybrid initialization method to achieve a higher performance improvement, as the case on Isolet v.s. Lung_discrete.

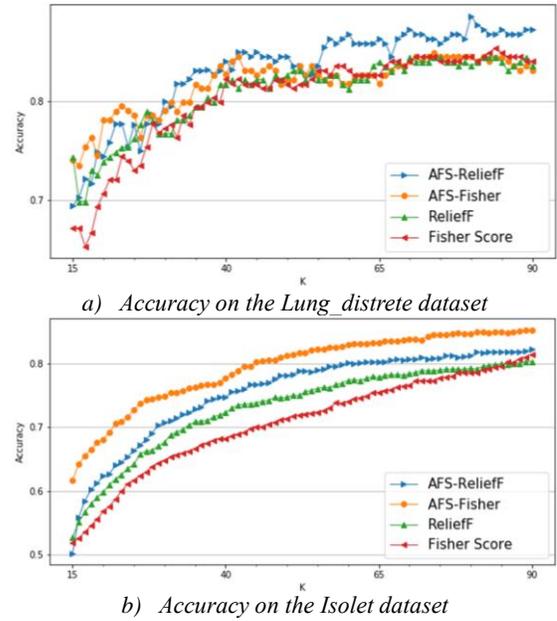

*a) Accuracy on the Lung_distrete dataset*

*b) Accuracy on the Isolet dataset*

*Figure 5. Accuracy with(out) the hybrid initialization methods, SVM-linear as classifier, K is the number of features selected*

## Conclusion

In this paper, a novel feature selection architecture is introduced that extends the attention mechanism to the general feature selection tasks. Specifically, by formulating the feature selection problem into a binary classification problem for each feature, we are able to identify each feature' weight according to its feature selection pattern. This architecture is designed to be detachable and allow either the attention module or the learning module to be trained individually. Thus, the reuse of learning model and the hybrid initialization become possible. Experiment results show that AFS can achieve the best feature selection accuracy on several different datasets. Results also demonstrate that AFS achieves the best feature selection stability in response to several different noises introduced in the n-MNIST variants. In future work, we aim to develop more domain-optimized solutions for specific structural data. We are also working on the structure optimization to reduce its computation cost on ultra-high dimensional datasets.

## Acknowledgments

This work is supported by the National Science Foundation of China (No. 61772473).

## Reference

Alelyani, S., H. Liu and L. Wang, 2011. The effect of the characteristics of the dataset on the selection stability. In: 2011 23rd IEEE International Conference on Tools with Artificial Intelligence. IEEE: pp: 970-977.

Bengio, Y., O. Delalleau, N. Roux, J.-F. Paiement, P. Vincent and M. Ouimet, 2003. Feature extraction: Foundations and applications, chapter spectral dimensionality reduction. Springer.

Bolón-Canedo, V., N. Sánchez-Maroño and A. Alonso-Betanzos, 2015. Recent advances and emerging challenges of feature selection in the context of big data. Knowledge-Based Systems, 86: 33-45.

Duda, R.O., P.E. Hart and D.G. Stork, 2012. Pattern classification. John Wiley & Sons.

Erhan, D., C. Szegedy, A. Toshev and D. Anguelov, 2014. Scalable object detection using deep neural networks. In: Proceedings of the IEEE Conference on Computer Vision and Pattern Recognition. pp: 2147-2154.

Girshick, R., J. Donahue, T. Darrell and J. Malik, 2014. Rich feature hierarchies for accurate object detection and semantic segmentation. In: Proceedings of the IEEE conference on computer vision and pattern recognition. pp: 580-587.

Grgic-Hlaca, N., M.B. Zafar, K.P. Gummadi and A. Weller, 2018. Beyond distributive fairness in algorithmic decision making: Feature selection for procedurally fair learning. In: Proceedings of the Thirty-Second AAAI Conference on Artificial Intelligence, New Orleans, Louisiana, USA.

Gui, J., Z. Sun, S. Ji, D. Tao and T. Tan, 2017. Feature selection based on structured sparsity: A comprehensive study. IEEE transactions on neural networks and learning systems, 28(7): 1490-1507.

He, X., D. Cai and P. Niyogi, 2006. Laplacian score for feature selection. In: Advances in neural information processing systems. pp: 507-514.

Jaderberg, M., K. Simonyan and A. Zisserman, 2015. Spatial transformer networks. In: Advances in neural information processing systems. pp: 2017-2025.

Kononenko, I., 1994. Estimating attributes: Analysis and extensions of relief. In: European conference on machine learning. Springer: pp: 171-182.

Laskar, Z. and J. Kannala, 2017. Context aware query image representation for particular object retrieval. In: Scandinavian Conference on Image Analysis. Springer: pp: 88-99.

Li, Y., C.-Y. Chen and W.W. Wasserman, 2015. Deep feature selection: Theory and application to identify enhancers and promoters. In: International Conference on Research in Computational Molecular Biology. Springer: pp: 205-217.

Liu, H., H. Mao and Y. Fu, 2016. Robust multi-view feature selection. In: Data Mining (ICDM), 2016 IEEE 16th International Conference on. IEEE: pp: 281-290.

Liu, J., S. Ji and J. Ye, 2009. Multi-task feature learning via efficient l 2, 1-norm minimization. In: Proceedings of the twenty-fifth conference on uncertainty in artificial intelligence. AUAI Press: pp: 339-348.

Ma, F., R. Chitta, J. Zhou, Q. You, T. Sun and J. Gao, 2017. Dipole: Diagnosis prediction in healthcare via attention-based bidirectional recurrent neural networks. In: Proceedings of the 23rd ACM SIGKDD International Conference on Knowledge Discovery and Data Mining. ACM: pp: 1903-1911.

Nie, F., H. Huang, X. Cai and C.H. Ding, 2010. Efficient and robust feature selection via joint ℓ2, 1-norms minimization. In: Advances in neural information processing systems. pp: 1813-1821.

Nie, F., S. Xiang, Y. Jia, C. Zhang and S. Yan, 2008. Trace ratio criterion for feature selection. In: AAAI. pp: 671-676.

Peng, H., F. Long and C. Ding, 2005. Feature selection based on mutual information criteria of max-dependency, max-relevance, and min-redundancy. IEEE Transactions on pattern analysis and machine intelligence, 27(8): 1226-1238.

Robnik-Šikonja, M. and I. Kononenko, 2003. Theoretical and empirical analysis of relieff and rrelieff. Machine learning, 53(1-2): 23-69.

Roy, D., K.S.R. Murty and C.K. Mohan, 2015. Feature selection using deep neural networks. In: Neural Networks (IJCNN), 2015 International Joint Conference on. IEEE: pp: 1-6.

Shumway, R., 1987. Statistics and data analysis in geology. Taylor & Francis.

Tang, J., S. Alelyani and H. Liu, 2014. Feature selection for classification: A review. Data classification: Algorithms and applications: 37.

Tong, B., M. Klinkigt, M. Iwayama, T. Yanase, Y. Kobayashi, A. Sahu and R. Vennelakanti, 2017. Learning to generate rock descriptions from multivariate well logs with hierarchical attention. In: Proceedings of the 23rd ACM SIGKDD International Conference on Knowledge Discovery and Data Mining. ACM: pp: 2031-2040.

Wang, Q., J. Zhang, S. Song and Z. Zhang, 2014. Attentional neural network: Feature selection using cognitive feedback. In: Advances in Neural Information Processing Systems. pp: 2033-2041.

Xu, K., J. Ba, R. Kiros, K. Cho, A. Courville, R. Salakhutdinov, R. Zemel and Y. Bengio, 2015. Show, attend and tell: Neural image caption generation with visual attention. Computer Science: 2048-2057.

Yao, L., A. Torabi, K. Cho, N. Ballas, C. Pal, H. Larochelle and A. Courville, 2015. Describing videos by exploiting temporal structure. In: Proceedings of the IEEE international conference on computer vision. pp: 4507-4515.

Yin, S., S.X. Ding, X. Xie and H. Luo, 2014. A review on basic data-driven approaches for industrial process monitoring. IEEE Transactions on Industrial Electronics, 61(11): 6418-6428.

Yin, W., H. Schütze, B. Xiang and B. Zhou, 2015. Abcnn: Attention-based convolutional neural network for modeling sentence pairs. arXiv preprint arXiv:1512.05193.

Zhang, Q., P. Zhang, G. Long, W. Ding, C. Zhang and X. Wu, 2015. Towards mining trapezoidal data streams. In: Data Mining (ICDM), 2015 IEEE International Conference on. IEEE: pp: 1111-1116.

Zhao, L., Q. Hu and W. Wang, 2015. Heterogeneous feature selection with multi-modal deep neural networks and sparse group lasso. IEEE Transactions on Multimedia, 17(11): 1936-1948.

Zhao, Z. and H. Liu, 2007. Spectral feature selection for supervised and unsupervised learning. In: Proceedings of the 24th international conference on Machine learning. ACM: pp: 1151-1157.